\documentclass[conference]{IEEEtran}
\IEEEoverridecommandlockouts
\usepackage{cite}
\usepackage{amsmath,amssymb,amsfonts}
\usepackage{algorithmic}
\usepackage{graphicx}
\usepackage{textcomp}
\usepackage{xcolor}
\usepackage{float} 
\usepackage{placeins}
\usepackage{caption}
\usepackage{subcaption}
\def\BibTeX{{\rm B\kern-.05em{\sc i\kern-.025em b}\kern-.08em
    T\kern-.1667em\lower.7ex\hbox{E}\kern-.125emX}}
\begin{document}

\title{Deep Reinforcement Learning with Enhanced PPO for Safe Mobile Robot Navigation\\
}

\author{\IEEEauthorblockN{Hamid Taheri}
\IEEEauthorblockA{\textit{KN Toosi University of Technology} \\
Tehran, Iran \\
hamidtaheri@email.kntu.ac.ir}

\and
\IEEEauthorblockN{Seyed Rasoul Hosseini}
\IEEEauthorblockA{\textit{KN Toosi University of Technology} \\
Tehran, Iran \\
rasoul.hosseini7@gmail.com}
\and
\IEEEauthorblockN{Mohammad Ali Nekoui}
\IEEEauthorblockA{\textit{KN Toosi University of Technology} \\
Tehran, Iran \\
manekoui@eetd.kntu.ac.ir
}
\and

}

\maketitle

\begin{abstract}
 Collision-free motion is essential for mobile robots. Most approaches to collision-free and efficient navigation with wheeled robots require parameter tuning by experts to obtain good navigation behavior. This study investigates the application of deep reinforcement learning to train a mobile robot for autonomous navigation in a complex environment. The robot utilizes LiDAR sensor data and a deep neural network to generate control signals guiding it toward a specified target while avoiding obstacles. We employ two reinforcement learning algorithms in the Gazebo simulation environment: Deep Deterministic Policy Gradient and proximal policy optimization.
The study introduces an enhanced neural network structure in the Proximal Policy Optimization algorithm to boost performance, accompanied by a well-designed reward function to improve algorithm efficacy. Experimental results conducted in both obstacle and obstacle-free environments underscore the effectiveness of the proposed approach. This research significantly contributes to the advancement of autonomous robotics in complex environments through the application of deep reinforcement learning.
\end{abstract}


\section{Introduction}
In an era where robotics is making profound strides in reshaping our world, the ability of robots to autonomously navigate complex and dynamic environments remains a challenge. Conventional robotic navigation systems often rely on meticulously crafted maps, limiting their adaptability to real-world scenarios where environments are constantly changing \cite{cadena2016}. This paradigm shift, marking a departure from conventional navigation methods, is made possible through the advanced capabilities of Deep Reinforcement Learning (DRL), wherein robots dynamically learn and adapt their navigation strategies based on real-time interactions with the environment \cite{arulkumaran2017}. This innovative approach not only enables autonomous exploration in novel surroundings but also signifies a transformative leap towards adaptive, intelligent robotic systems capable of navigating diverse and unpredictable landscapes \cite{tai2016}. Additionally, the emergence of deep learning has led to significant progress in autonomous systems, including advancements in lane detection for self-driving cars, which leverage machine learning techniques to handle dynamic scenarios\cite{hosseini2024}.

\begin{figure}[htbp]
\centering
\includegraphics[height=1.2in]{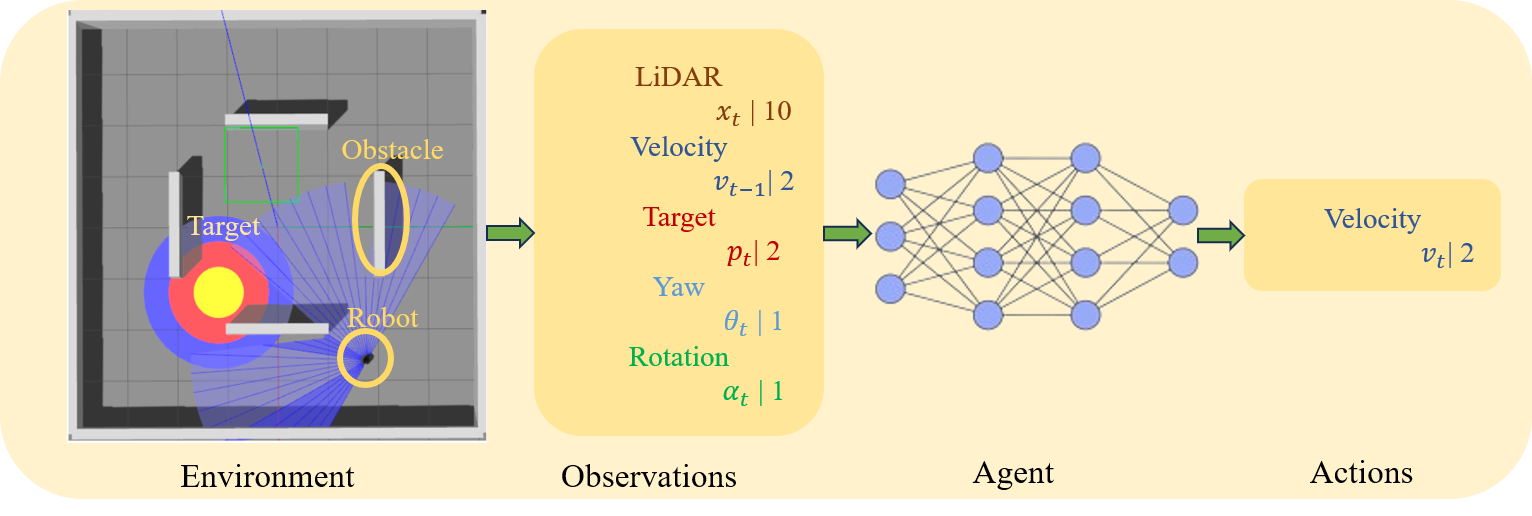}
\caption{A mapless motion planner was trained using the Proximal Policy Optimization (PPO) algorithm to guide a nonholonomic mobile robot to its target position while avoiding collisions.}
\label{fig:diagram}
\end{figure}

Mapless navigation represents a significant departure from conventional navigation strategies that heavily depend on static, pre-built maps. Instead, it empowers robots with the capability to explore and maneuver through uncharted territories, responding dynamically to unforeseen obstacles, and adapting to evolving circumstances \cite{cadena2016}. This paradigm shift holds immense promise in a wide range of applications, from search and rescue missions in disaster-stricken areas to autonomous exploration of remote and hazardous locations \cite{tai2016}.

Deep Reinforcement Learning (DRL), a subset of machine learning, has become a powerful tool for enhancing robots' navigation skills through experiential learning \cite{sutton2018}. By employing DRL algorithms, robots can make informed decisions in real-time, learning from their actions and refining their path-planning strategies to meet specific objectives. This adaptive learning process mirrors human-like decision-making, enabling robots to navigate smoothly and effectively even in the absence of conventional maps.

The Robot Operating System (ROS) has established itself as the predominant standard for developing and controlling robotic systems. Offering a versatile framework, ROS facilitates the seamless integration of diverse sensors, actuaries, and control algorithms \cite{quigley2009}. When coupled with Gazebo, a high-fidelity robot simulation environment, ROS allows for rigorous testing and validation of mapless navigation algorithms in a secure and controlled virtual setting before their deployment in real-world scenarios \cite{koenig2004}.

This paper delves into the intriguing realm of mapless robot navigation using Deep Reinforcement Learning (DRL). We explore the theoretical foundations of DRL and its practical applications, demonstrating how robots can navigate without relying on pre-existing maps. Building on this foundation, we present an innovative approach to mapless navigation, marked by three key contributions.

Firstly, we introduce a Comprehensive Observation Setup that encompasses a wide array of inputs to fully capture the robot's environment. These inputs include laser readings across 30 dimensions, the robot's past linear and angular velocities, and the target position relative to the robot, expressed in polar coordinates. Additionally, we consider the robot's yaw angle and the orientation required to face the target. This multidimensional observation framework facilitates a comprehensive understanding of the robot's surroundings, empowering the navigation system with enhanced decision-making capabilities.

PPO-Based Learning for Navigation: In our strategy for training the motion planner, we introduce a customized neural network architecture designed specifically for the Proximal Policy Optimization (PPO) algorithm. This modification results in a significant improvement in the overall performance of the planner. Leveraging this reinforcement learning technique, the planner becomes proficient in acquiring effective navigation strategies by optimizing its decision-making process based on available sensory information. Notably, the planner's capability to directly output continuous linear and angular velocities contributes to a streamlined and efficient navigation process.

Through these contributions, we advance the ability of PPO algorithm, offering an innovative solution that combines sparse sensor data, reinforcement learning with PPO, and adaptability to simulated scenarios, promising efficient and adaptable navigation capabilities for robots in a static environments within the simulation context.

\section{Related Work}

Mobile robot navigation has been a significant research area in robotics, and the integration of deep reinforcement learning (DRL) methods for continuous control has gained substantial attention in recent years. This section provides an overview of relevant literature and highlights key contributions in the field.

\subsection{Navigational Techniques in Robotics}
Traditional methods for mobile robot navigation often rely on techniques such as Simultaneous Localization and Mapping (SLAM) and path planning algorithms like A* and Dijkstra's. While these methods have proven effective in structured environments, they may struggle in dynamic, unstructured, or partially observable settings.
The application of reinforcement learning in robotics has opened new avenues for autonomous navigation. Early approaches focused on discrete action spaces and tabular methods. Researchers successfully applied RL algorithms to tasks like maze navigation and grid world problems. However, the discrete nature of actions limited their applicability to real-world scenarios, especially in continuous control settings.
Deep reinforcement learning introduced the use of neural networks to approximate Q-functions or policy functions, enabling robotic agents to handle high-dimensional state and action spaces. Prominent DRL algorithms like Deep Q-Networks (DQN) \cite{dqn_paper} and Trust Region Policy Optimization (TRPO) \cite{trpo_paper} showed promising results in challenging environments, but their application to mobile robot navigation remained limited due to the need for discrete action spaces.
The breakthrough in continuous control came with algorithms like the Deep Deterministic Policy Gradient (DDPG)\cite{ddpg_paper} and Proximal Policy Optimization (PPO) \cite{b5}. These methods facilitated mobile robot navigation in real-world scenarios, enabling smooth and precise control. PPO and DDPG, in particular, allowed robots to learn continuous actions, making it a significant milestone in this field.

\subsection{Prior Work in Mobile Robot Navigation}
Several studies have already applied Deep Reinforcement Learning (DRL) to mobile robot navigation, demonstrating the method's adaptability and potential in a variety of contexts.

Researchers have made significant strides in the field of DRL for robotic navigation. Silver et al. \cite{b1} utilized the Deep Deterministic Policy Gradient (DDPG) algorithm to empower mobile robots to autonomously navigate through cluttered environments while effectively avoiding obstacles. This study showcases the potential of DRL for enhancing obstacle avoidance capabilities in robots. Zhu et al. \cite{b2} pioneered end-to-end navigation by combining convolutional neural networks (CNNs) with DRL. This integration allows robots to process visual data for perception and make navigation decisions dynamically, thus enabling autonomous navigation without explicit programming. The approach exemplifies how DRL can be integrated with other machine learning techniques to improve the autonomy of robotic systems.

Tobin et al. \cite{b3} explored the challenges of Sim-to-Real Transfer, where robots are trained in simulated environments with the aim of applying these learned policies to real-world scenarios. Their work addresses key deployment challenges and highlights the effectiveness of simulation as a training ground for real-life applications. Furthermore, Zheng et al. \cite{b4} investigated multi-agent navigation, utilizing multi-agent reinforcement learning to enable teams of robots to collaboratively navigate complex environments. Their research demonstrates how DRL can be extended to coordinate multiple agents, enhancing the collective navigation capabilities in scenarios like search and rescue or synchronized surveying.

These studies collectively underscore the versatility and expansive potential of DRL in advancing the field of robotic navigation, paving the way for more sophisticated and autonomous robotic systems.

\section{Methodology}

\subsection{Proximal Policy Optimization (PPO)}

In our research, we employed the Proximal Policy Optimization (PPO) algorithm, a well-regarded reinforcement learning technique known for its stability and efficacy in continuous control tasks. Unlike the Deep Deterministic Policy Gradient (DDPG) algorithm, PPO iteratively updates the policy to maximize the expected cumulative reward, implementing a constraint to limit policy changes and prevent large, abrupt deviations. This controlled approach to policy updates ensures more stable training progress, making PPO particularly suitable for our mapless motion planning problem.

In the Proximal Policy Optimization (PPO) algorithm, the loss function is composed of two integral components: the policy loss and the value loss.

The policy loss is crafted to modulate the updates made to the policy, ensuring they remain modest and do not significantly diverge from the current policy. This controlled adjustment is vital for the stability of the training process, enabling a gradual and steady improvement in policy performance. Mathematically, the policy loss can be described as follows:

\begin{equation}
\mathcal{L}^{\text{clip}}(\theta) = \mathbb{\hat E}\left[ \min \left( r_t (\theta) \mathcal{\hat A}, \text{clip}(r_t (\theta), \epsilon+1, \epsilon-1) \right) \right]
\end{equation}

\begin{equation}
r_t(\theta) = \frac{\pi_\theta(s_t, a_t)}{\pi_{\theta_{old}}(s_t, a_t)},\hspace{5mm} \epsilon=0.2
\end{equation}

Here $\mathcal{L}^{\text{clip}}(\theta)$ represents the clipped surrogate objective, $\theta$ denotes the parameters of the policy, $r_t(\theta)$ is the probability ratio of the new policy to the old policy, $\mathcal{\hat A}$ is the advantage function, and $\epsilon$ is a hyperparameter determining the extent of allowed policy changes.

The value loss is associated with the critic network and is aimed at minimizing the difference between the predicted value $V(s_t)$ and the actual discounted cumulative reward $R_t$. The value loss is calculated as:

\begin{equation}
L^{VF}(\theta) = \mathbb{E} \left[ \left( V(s_t) - R_t \right)^2 \right]
\end{equation}

In our implementation of the Proximal Policy Optimization (PPO) algorithm, the actor and critic networks are fundamental components that facilitate the learning process. The actor network is responsible for defining the policy, which specifies the probability distribution of possible actions in a given state. Conversely, the critic network provides an estimate of the value of each state, essentially predicting the expected return from that state.

The training of the actor network is guided by the policy loss, which helps in refining the policy to ensure better decision-making in navigating the environment. On the other hand, the critic network is trained to minimize the value loss, aiming to enhance the accuracy of state value predictions.

This dual-network approach, where the actor and critic are concurrently trained, is central to the efficiency of the PPO algorithm. By iteratively updating both networks, our system continuously improves in its ability to not only choose optimal actions (via the actor) but also in evaluating the potential future rewards of current states (via the critic). This collaborative training mechanism is particularly effective in our application of mapless motion planning, allowing for a more effective navigation strategy to be developed by the algorithm.

\subsection{Problem Definition}

In this study, we confront the challenge of mapless motion planning for mobile ground robots, with the primary objective to develop a robust translation function for determining the next velocity $v_{t}$ of a robot based on its current state $s_{t}$. The state $s_{t}$ encompasses several critical components, formalized as:

\begin{equation}
v_t = f(x_t, p_t, v_{t-1}, \theta_t, \alpha_t)
\end{equation}

Where Sensor Information $x_{t}$ consists of data from the robot's sensors to understand the environment, and Relative Target Position $p_{t}$ denotes the location of the target relative to the robot. The Previous Velocity $v_{t-1}$ indicates the robot's last recorded speed, aiding in dynamic stability. The Yaw Angle $\theta_{t}$ specifies the robot's orientation, and the Rotation Degree $\alpha_t$ is crucial for aligning the robot with the target.

Our goal is to model these states into actionable insights, specifically to compute the next velocity $v_{t}$, facilitating agile and accurate navigation in dynamic settings.

\subsection{Data Processing}

This section details our LiDAR data processing approach to enhance robot navigation in a complex environment. We aim to condense raw sensor data for optimal decision-making by Actor-Critic models while minimizing computational load.

The LiDAR sensor captures 30 distance measurements of the robot's surroundings. To align with Actor-Critic architecture and ensure efficiency, we propose dividing these measurements into 10 batches of three data points each. Within each batch, we select the minimum distance, identifying the closest obstacle in the robot's field of view. This process yields 10 streamlined observations.

This approach enhances the efficiency and effectiveness of our navigation models. By focusing on the nearest obstacle in each batch, we reduce data dimensionality, streamline computational processes, and prioritize relevant information for safe navigation. These selected data points provide actionable insights for generating well-informed navigation commands, optimizing our robotic systems' navigation capabilities.

\begin{figure}
\centering
\includegraphics[height=1.8in]{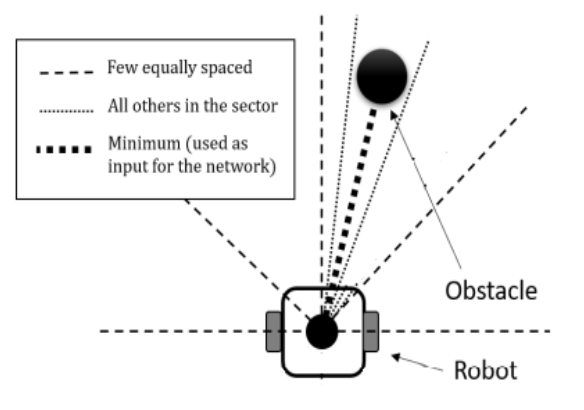}
\caption{Distilling raw sensor data into concise observations for optimal decision-making.}
\label{fig:sensor_data}
\end{figure}

As shown in Figure \ref{fig:sensor_data}, the LiDAR data processing approach identifies the nearest obstacle within each batch, distilling raw sensor data into concise observations for optimal decision-making in robot navigation.

\subsection{Model Architecture}

In the study, we leverage the Proximal Policy Optimization (PPO) algorithm to cultivate our tailored model for efficient navigation.

\begin{figure}[htbp]
\centering
\includegraphics[height=2.1in]{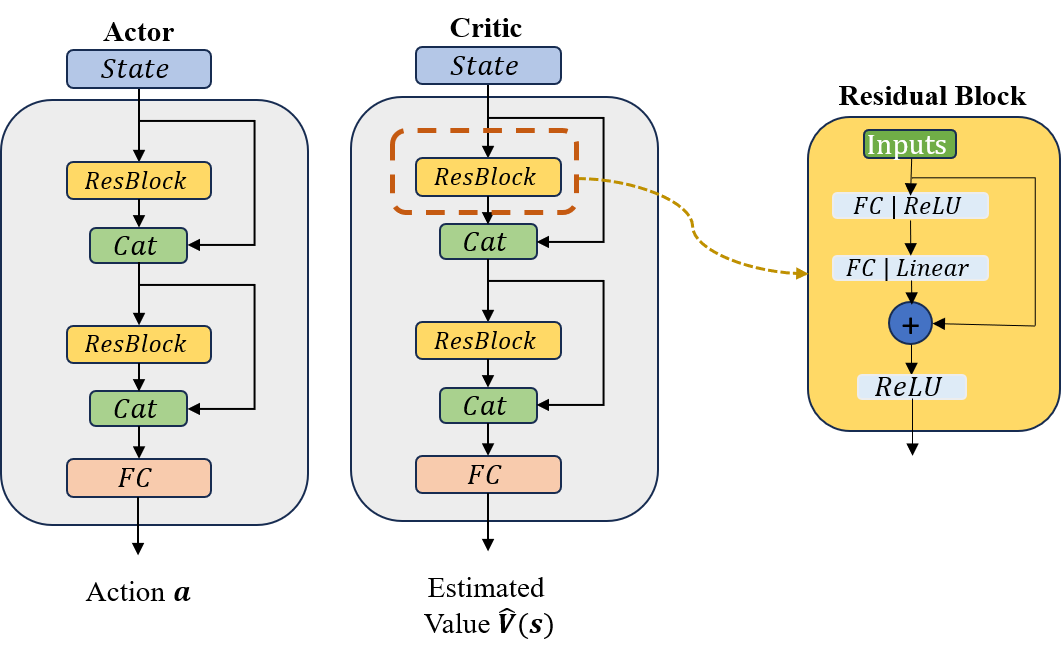}
\caption{Neural Network Architecture of the PPO Algorithm illustrating layer
types, dimensions, and activations, with a merged output in the Merge layer.}
\label{fig:model_Architecture}
\end{figure}

A shown in figure \ref{fig:model_Architecture}, In our innovative approach, we introduced an advanced architecture within the actor and critic networks of our Proximal Policy Optimization (PPO) model, significantly enhancing its performance. Central to this enhancement are the Residual Blocks (ResBlocks) integrated into both networks.

Our system operates with a 16-dimensional observation space designed to meticulously capture environmental nuances. The action space, comprising linear and angular velocities, is adeptly constrained to mirror realistic robotic maneuvers; angular velocity is confined to a range of (-1, 1) using the hyperbolic tangent function ($\tanh$), while linear velocity is restricted to (0, 1) via a sigmoid function, accommodating the robot’s limited reverse capability due to sparse rear sensor coverage.

For environmental perception, we process laser range data, sampling it uniformly from -90 to 90 degrees and normalizing these readings to a (0, 1) scale. This data handling facilitates a more structured and effective decision-making process.

To align with the dynamics of the Turtlebot, we set the maximum speeds at 0.25 m/s (linear) and 1 rad/s (angular). The integration of two ResBlocks in each of the actor and critic networks is a pivotal enhancement. These blocks significantly improve the flow of information and the efficiency of the control policy, proving indispensable for the robot's adept navigation in environments without predefined maps.

\begin{table}[htbp]
\caption{PPO Critic network structure details}
\centering
\begin{tabular}{|p{2cm}|p{2cm}|p{2cm}|}
\hline
\textbf{ } & \textbf{Weight Size} & \textbf{Activation}  \\
\hline
1st layer & 512 &  LeakyRelu \\
2nd layer & 16  & Linear \\
res & input &  - \\
Concatenate 1 & input &  LeakyRelu \\
3rd layer & 512 &  LeakyRelu \\
4th layer & 32 &  LeakyRelu \\
res & Concatenate 1 &  - \\
Concatenate 2 & Concatenate 1 &  LeakyRelu \\
\hline
5th layer & 1 &  Linear \\
\hline
lr & \multicolumn{2}{|c|}{3e-4} \\
\hline
\end{tabular}
\label{tab:critic}
\end{table}

\begin{table}[htbp]
\caption{PPO Critic network structure details}
\centering
\begin{tabular}{|p{2cm}|p{2cm}|p{2cm}|}
\hline
\textbf{ } & \textbf{Weight Size} & \textbf{Activation}  \\
\hline
1st layer & 512 &  LeakyRelu \\
2nd layer & 16  & Linear \\
res & input &  - \\
Concatenate 1 & input &  LeakyRelu \\
3rd layer & 512 &  LeakyRelu \\
4th layer & 32 &  LeakyRelu \\
res & Concatenate 1 &  - \\
Concatenate 2 & Concatenate 1 &  LeakyRelu \\
\hline
\vspace{0.03cm} 5th layer & \vspace{0.03cm} 2 &  Sigmoid\newline Tanh \\
\hline
lr & \multicolumn{2}{|c|}{3e-4} \\
\hline
\end{tabular}
\label{tab:actor}
\end{table}

The ResBlocks, by enabling shortcut connections within the networks, facilitate the gradient flow during training, thus mitigating the vanishing gradient problem and leading to more stable and faster learning. This architecture allows the critic network to more accurately estimate the state's value (V-value), with a linear activation function employed in the output layer for a clear, interpretable signal.

This architectural innovation underpins our model’s ability to learn and execute mapless navigation tasks with increased efficiency and effectiveness, demonstrating a significant advancement in autonomous robotic navigation technology.

\subsection{Reward Function}

The reward function is crucial in the reinforcement learning process, acting as a guide by rewarding desirable actions and penalizing unfavorable ones. The agent aims to develop a policy that maximizes these rewards, optimizing decision-making to fulfill its objectives in the given environment. The construction of the reward function significantly dictates the agent's behavior and learning effectiveness.

Our reward function, expressed mathematically, rewards the agent for approaching the target, penalizes potential collisions, and incentivizes progress towards the target:

\begin{equation}
r_1(s_t, a_t) = \left\{
\begin{array}{@{}ll}
r_{\text{arrive}} & \text{if } d_t < c_d\\
r_{\text{collision}} & \text{if } \max_{x_t} < c_o \\
c_r(d_{t-1} - d_t) & \text{otherwise}.
\end{array}
\right.
\label{eq:first_reward}
\end{equation}

In equation~\ref{eq:first_reward}, $r_{\text{arrive}}$ is granted when the agent is within a critical distance $c_d$ to the target, promoting goal-oriented movement and $r_{\text{collision}}$ is imposed if any sensor reading $x_t$ signals a near-collision distance $c_o$, promoting safety.
Otherwise, the reward is proportional to the reduction in distance to the target compared to the last timestep, encouraging consistent progress.
This structure drives the agent towards the target while avoiding hazards, balancing goal achievement, safety, and efficient navigation.
While this reward function is a cornerstone of our reinforcement learning agent's behavior, its applicability might be limited in obstacle-rich environments. The balance it strikes between target-seeking behavior, safety, and steady advancement may need adjustment or augmentation to better suit scenarios where obstacles densely populate


the environment (see Equation \ref{eq:second_reward}). Consideration of alternative reward structures may be necessary to ensure effective learning and decision-making in such challenging settings. Consequently, we have designed a new reward function that penalizes the robot as it moves toward the walls and rewards the agent exponentially as it approaches the target, aiming to address the challenges posed by obstacle-filled environments.

\begin{equation}
r_2(s_t, a_t) = \left\{
\begin{array}{@{}ll}
r_{\text{arrive}} & \text{if } d_t < c_d \\
r_{\text{collision}} & \text{if } x_t < c_o \\
c_r (d_{t-1} - d_t) \times 2^{\left( \frac{d_{t-1}}{d_t} \right)} - c_p \left( 1 - hd \right) & \text{otherwise}.
\end{array}
\right.
\label{eq:second_reward}
\end{equation}

Where \( hd \) represents the heading deviation of the sensor, \( c_o \) represents the collision threshold, \( c_d \) represents the target proximity threshold, \( c_r \) represents the reward coefficient, and \( c_p \) represents the penalty coefficient.

\section{Simulation}
\subsection{Environmental Setup}

The training of our model was conducted in virtual environments, using the Robot Operating System (ROS) combined with the Gazebo simulator. These platforms provided a realistic and customizable setting for the experiments.

\begin{figure}[H]
\centering
\includegraphics[height=1.5in]{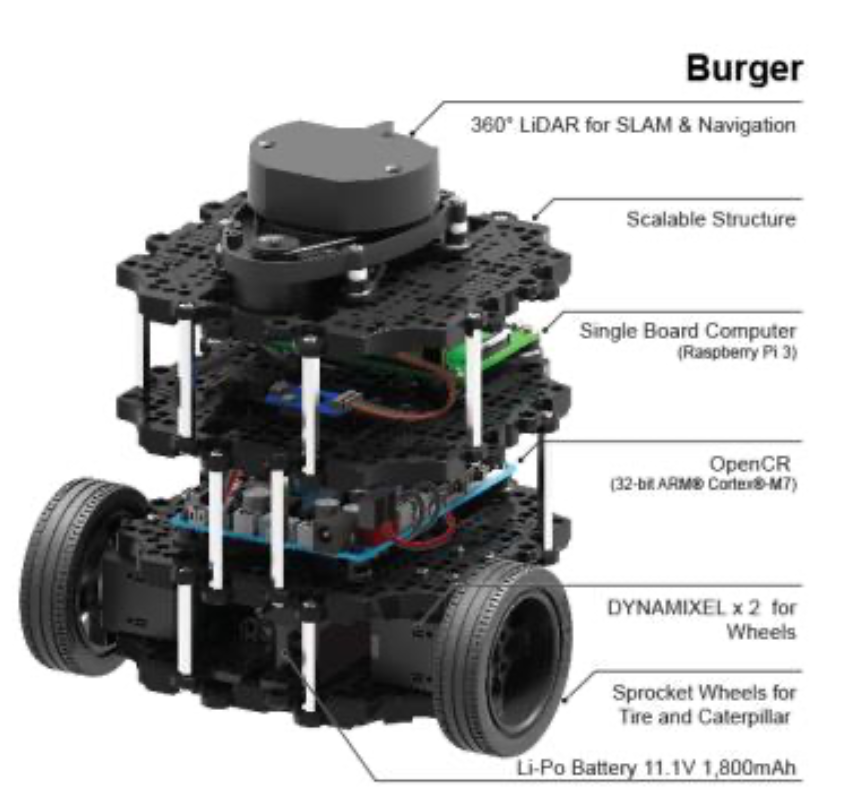}
\caption{The structure of Turtlebot3 robot.}
\label{fig:Robot}
\end{figure}

\begin{figure*} 
\centering
\begin{subfigure}{0.4\textwidth} 
    \centering 
    \includegraphics[width=0.4\linewidth]{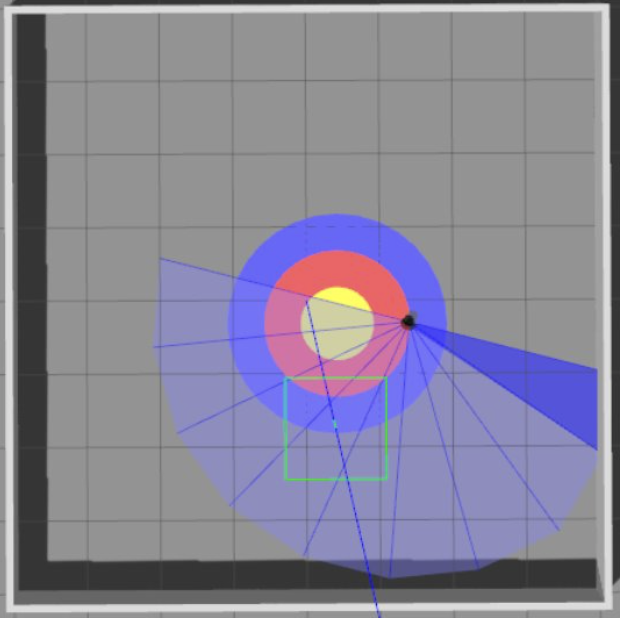} 
    \caption{The simple environment}
    \label{fig:env1}
\end{subfigure}%
\hfill
\begin{subfigure}{0.4\textwidth} 
    \centering 
    \includegraphics[width=0.4\linewidth]{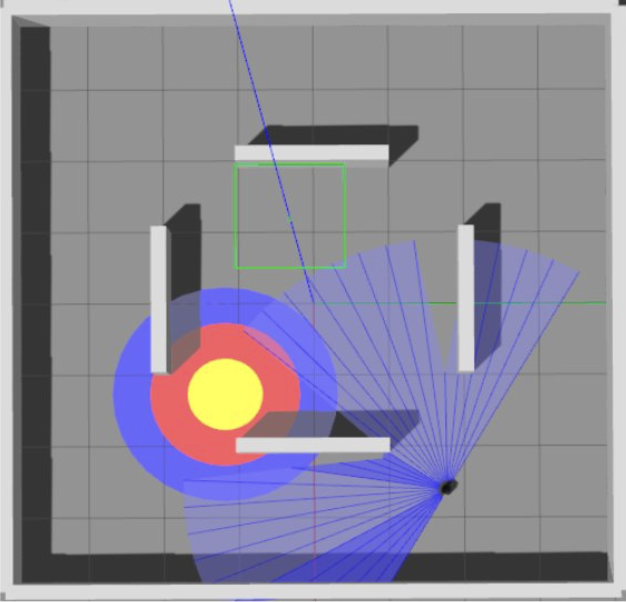} 
    \caption{The complex environment}
    \label{fig:env2}
\end{subfigure}
\caption{Comparative view of two different robotic environments}
\label{fig:both-envs}
\end{figure*}

A shown in figure \ref{fig:both-envs}, We conducted our experiments in two distinct environments: an obstacle-free environment and a complex environment. Both environments were simulated indoors within a 10 x 10 square meter area, enclosed by walls. The complex environment was additionally populated with obstacles strategically placed to challenge the navigation capabilities of our robot. Throughout the experiments, we utilized the Turtlebot as the robotic platform for these trials (Figure \ref{fig:Robot}).

An important aspect of our setup was the representation of the target as a cylindrical object. However, the Turtlebot's laser sensor was inherently unable to detect this object directly. This limitation is critical for understanding the robot’s perception and navigational abilities within the environment.

\begin{figure}[htbp]
\centering
\includegraphics[height=2.4in]{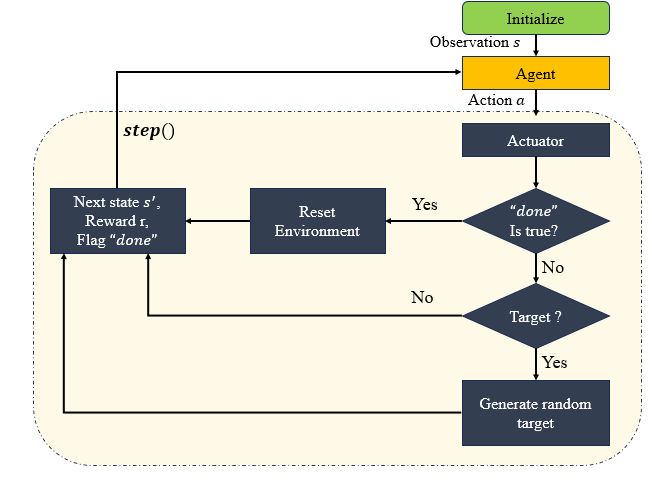}
\caption{The flowchart of partial DRL environment.}
\label{fig:Flowchart}
\end{figure}

In Figure \ref{fig:Flowchart}, the agent interacts with the Gazebo environment. Following initialization, the agent selects an action $a$ to interact with the environment. As the agent progresses, it checks for collisions or reaching the maximum time step, indicated by the Boolean flag \texttt{done}. If \texttt{done} is true, the environment resets, providing the current reward $r^*$, the next state $s'$, and the \texttt{done} flag. 

If \texttt{done} is false, and the agent has reached the target position, another target position is generated, returning $(r', s', \texttt{done})$. If the target position has not been reached, it returns $(r'', s', \texttt{done})$ directly. Here, $r^*$/$r'$/ $r''$ denote the reward $r_t$ under different conditions. This process continues until the maximum time step is reached, with the collected data $(s, a, r, s', \texttt{done})$ utilized for DRL algorithm training.

For each training episode, the target's starting position was randomized within the environment, ensuring it was placed away from obstacles to prevent immediate collisions. This approach of random initialization was vital for training the model to adapt to a wide range of navigation scenarios, enhancing its ability to operate effectively in varied environments.

\subsection{Performance Comparison}



In our study, we explore the efficacy of Proximal Policy Optimization (PPO) enhanced with Residual Blocks across different environmental complexities and reward function designs. We benchmark these against the Deep Deterministic Policy Gradient (DDPG) algorithm to evaluate performance variations. The investigation unfolds through three distinct scenarios:

In this scenario, we aim to showcase the efficacy of ResBlock PPO by comparing it with DDPG and vanilla PPO. Thus, we've crafted a simplified environment with a basic reward function depicted in equation \ref{eq:first_reward}. This initial setup allows us to gauge the performance attributes of the ResBlock-enhanced PPO algorithm under fundamental conditions.

Complex environment and basic reward function outlined in equation \ref{eq:first_reward}: subsequently, we increase the environmental complexity to observe how effectively the ResBlock PPO setup, still utilizing the basic reward function, adapts to and navigates within more challenging contexts.

Complex environment and advanced reward function outlined in equation \ref{eq:second_reward}: in the final configuration, we introduce an advanced reward function to our complex environment scenario, aiming to discern the impact of intricate reward structuring on the ResBlock PPO's navigational efficacy.

\section{Experimental Results}
\subsection{Simple Environment with Basic Reward}
In the first scenario, the ResBlock PPO algorithm demonstrated significant advantages over the DDPG method within the simple environment. Particularly, ResBlock PPO achieved faster navigation, indicating a more efficient and responsive decision-making process. This effectiveness is partly due to the ResBlock architecture, which helps maintain a strong gradient flow during training, essential for quick learning and adaptation.

\begin{figure}[H]
\centering
    \includegraphics[height=2.1in]{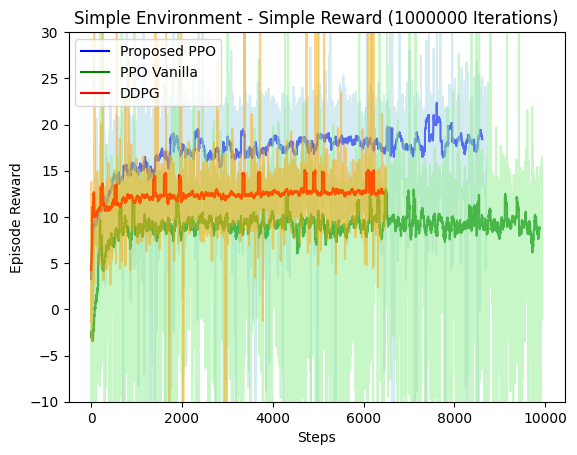}
\caption{Cumulative Reward of Proposed PPO and DDPG in a Simple Environment with Basic Reward Function.}
\label{fig:SESR}
\end{figure}

The improved performance depicted in Figure \ref{fig:SESR} shows that ResBlock PPO not only converges more rapidly but also adapts swiftly to environmental changes. This leads to quicker and more accurate pathfinding compared to both DDPG and vanilla PPO.

The architecture's ability to process environmental cues effectively allows for faster adjustments to changes, enhancing the robot's navigation capabilities. This trial establishes a clear performance ranking with ResBlock PPO at the forefront, followed by DDPG, and then vanilla PPO. The results underscore the potential of integrating advanced neural network architectures like ResBlock into reinforcement learning frameworks to boost efficiency and adaptability in robotic navigation.

\begin{table}[H]
\caption{Comparison of Proposed PPO vs DDPG}
\centering
\begin{tabular}{|p{1.5cm}|p{1.5cm}|p{1cm}|p{1cm}|p{1.5cm}|}
\hline
\textbf{Algorithm} & \textbf{Avg. Reward} & \textbf{Episodes} & \textbf{Success \%} & \textbf{Avg. Steps/Ep}  \\
\hline
\textbf{Prop. PPO} & \textbf{17.49} & \textbf{90} & \textbf{100} & \textbf{111.11} \\
\hline
Vanilla PPO & 3.34 & 23 & 30.43 & 419.69\\
\hline
DDPG & 12.65 & 65 & 98.46 & 144.92 \\
\hline
\end{tabular}
\label{tab:SESR}
\end{table}

\subsection{Complex Environment with Basic Reward}

In a more challenging environment characterized by intricate obstacles, the DDPG algorithm showcased superior navigation accuracy over the ResBlock PPO.

\begin{figure}[htbp]
\centering
\includegraphics[height=2.1in]{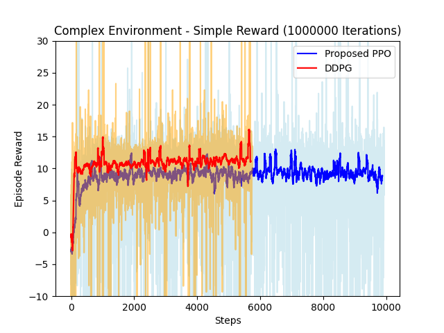}
\caption{Cumulative Reward of Proposed PPO and DDPG in a Complex Environment with Basic Reward Function.}
\label{fig:SRCE}
\end{figure}

\begin{figure*}
\centering
\begin{subfigure}[b]{0.45\textwidth}
    \centering
    \includegraphics[height=2in]{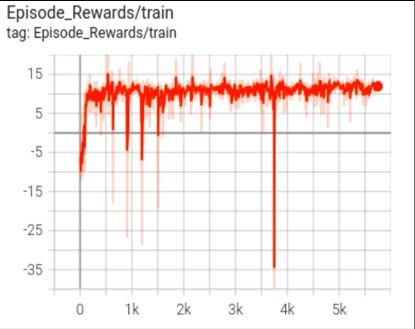}
    \caption{Cumulative Reward for Proposed DDPG Algorithm in a Complex Environment with Basic Reward Function}
    \label{fig:ddpg_reward}
\end{subfigure}%
\hfill
\begin{subfigure}[b]{0.45\textwidth}
    \centering
    \includegraphics[height=2in]{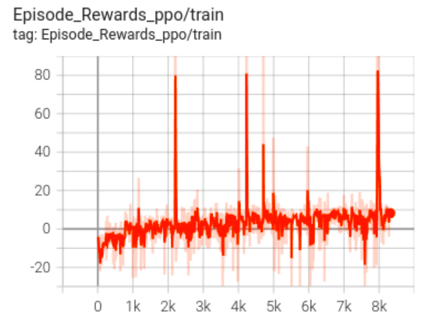}
    \caption{Cumulative Reward for Proposed PPO Algorithm in a Complex Environment with Advanced Reward Function}
    \label{fig:ppo_reward}
\end{subfigure}
\caption{Comparative analysis of cumulative rewards using DDPG and PPO algorithms}
\label{fig:both_rewards}
\end{figure*}
Despite encountering more complex spatial dynamics, DDPG demonstrated enhanced precision in navigation, underscoring its robustness in handling intricate obstacles as shown in \ref{tab:SRCE}. This performance discrepancy suggests that while ResBlock PPO excels in simpler scenarios, DDPG is better suited to environments requiring careful management of detailed spatial information and strategic maneuvering.

The results underscore a significant interplay between environmental complexity and algorithmic strengths, with DDPG particularly excelling in settings demanding high accuracy and strategic navigation maneuvers.

As depicted in Figure \ref{fig:SRCE}, the cumulative reward comparison graph provides a visual representation of the performance disparity between the proposed PPO and DDPG algorithms in a complex environment with a basic reward function.

This visualization corroborates our findings, further substantiating the efficacy of DDPG in environments necessitating precise navigation amidst intricate obstacles.

\begin{table}[H]
\caption{Proposed PPO vs DDPG - Complex Env}
\centering
\begin{tabular}{|p{1.5cm}|p{1.5cm}|p{1cm}|p{1cm}|p{1.5cm}|}
\hline
\textbf{Algorithm} & \textbf{Avg. Reward} & \textbf{Episodes} & \textbf{Success \%} & \textbf{Avg. Steps/Ep}  \\
\hline
\textbf{DDPG} & \textbf{11.23} & 63 & \textbf{93.65} & 158.73 \\
\hline
Prop. PPO & 9.07 & \textbf{87} & 67.81 & \textbf{114.94} \\
\hline
\end{tabular}
\label{tab:SRCE}
\end{table}


\subsection{Complex Environment with Advanced Reward}

In a more challenging environment, the application of an advanced reward function significantly enhanced the performance of ResBlock PPO. However, despite the improvement seen in PPO results due to the advanced reward function, DDPG still outperformed it in terms of success rate. Notably, while DDPG achieved a better success rate, it operated slower compared to PPO. This advanced reward function, tailored to assess finer details of the navigation strategy, better complemented ResBlock PPO’s adaptive learning capabilities. It effectively encouraged more strategic decision-making, allowing ResBlock PPO to navigate complex settings more proficiently than the previous version with a simple reward function. This performance improvement highlights the importance of aligning reward mechanisms with specific algorithm strengths, particularly in challenging environments where nuanced decision-making is crucial.

As shown in Figure \ref{fig:both_rewards} and Table \ref{tab:CRCE}, the comparative analysis of cumulative rewards using DDPG and PPO algorithms illustrates the efficacy of ResBlock PPO in environments with advanced reward functions.

\begin{table}[htbp]
\caption{Proposed PPO vs DDPG - Complex Env and new reward function}
\centering
\begin{tabular}{|p{1.5cm}|p{1.5cm}|p{1cm}|p{1.3cm}|p{1.5cm}|}
\hline
\textbf{Algorithm} & \textbf{Reward Function} &\textbf{Episodes} & \textbf{Success \%} & \textbf{Avg. Steps/Ep} \\
\hline
\textbf{DDPG} & benchmark & 63 & \textbf{93.65} & 158.73 \\
\hline
Prop. PPO  & Advanced &74 & 85.13 & 134.81 \\
\hline
Prop. PPO & Simple & 87 & 67.81 & 114.94 \\
\hline
\end{tabular}
\label{tab:CRCE}
\end{table}

\section{Challenges and Gaps}

The application of Deep Reinforcement Learning (DRL) in mobile robot navigation, despite its promise, faces several significant challenges that hinder its wider adoption and effectiveness.

A primary challenge is partial observability, where robots lack complete state information, leading to suboptimal decision-making. Enhancing sample efficiency is another critical issue, as DRL typically requires extensive interactions with the environment, which is impractical in real-world settings due to cost and risk. Safety in dynamic and unstructured environments remains a paramount concern. DRL systems must make safe decisions under uncertainty, handling edge cases without causing harm. The lack of robust benchmarking environments and standardized evaluation metrics further complicates the assessment of DRL algorithms, making it difficult to gauge their true efficacy and safety.

Addressing these challenges requires technical advancements and collaborative efforts among researchers to define rigorous testing and safety standards. Progress in these areas is essential for the reliable deployment of autonomous robots in real-world applications, such as industrial automation and search and rescue missions, where effective and safe navigation is critical.

\section{Conclusion}

This study demonstrated the successful application of deep reinforcement learning (DRL) to enhance autonomous navigation in mobile robots. Using advanced algorithms like Deep Deterministic Policy Gradient (DDPG) and enhanced Proximal Policy Optimization (PPO) in the Gazebo simulation environment, we trained robots to efficiently navigate complex environments.

By integrating LiDAR sensor data with DRL algorithms, robots were able to navigate towards targets while avoiding obstacles. Enhancements in the PPO neural network architecture and refined reward functions significantly boosted performance and training efficacy. Experimental results in various environments confirmed the effectiveness of our approach, highlighting the potential of DRL to advance autonomous robotics. This research contributes to the field by showing how improvements in algorithms and training can lead to substantial gains in robotic navigation, with promising applications in industrial, commercial, and rescue operations.

\vspace{12pt}



\begin{thebibliography}{00}

\bibitem{cadena2016} C. Cadena et al., "Past, present, and future of simultaneous localization and mapping: Toward the robust-perception age," IEEE Transactions on Robotics, vol. 32, no. 6, pp. 1309-1332, 2016.

\bibitem{arulkumaran2017} K. Arulkumaran et al., "Deep reinforcement learning: A brief survey," IEEE Signal Processing Magazine, vol. 34, no. 6, pp. 26-38, 2017.

\bibitem{tai2016} L. Tai and M. Liu, "Towards cognitive exploration through deep reinforcement learning for mobile robots," in IEEE/RSJ International Conference on Intelligent Robots and Systems (IROS), 2016.

\bibitem{hosseini2024} S. R. Hosseini and M. Teshnehlab, "ENet-21: An Optimized Light CNN Structure for Lane Detection," arXiv preprint arXiv:2403.19782, 2024.


\bibitem{sutton2018} R.S. Sutton, A.G. Barto, "Reinforcement learning: An introduction," MIT Press, 2018.
\bibitem{quigley2009} M. Quigley, K. Conley, B. Gerkey, J. Faust, T. Foote, J. Leibs, R. Wheeler, A. Y. Ng, "ROS: an open-source robot operating system," in ICRA Workshop on Open Source Software, Kobe, Japan, 2009.
\bibitem{koenig2004} N. Koenig, A. Howard, "Design and use paradigms for gazebo, an open-source multi-robot simulator," in Proceedings of the IEEE/RSJ International Conference on Intelligent Robots and Systems (IROS), 2004.

\bibitem{dqn_paper} V. Mnih, K. Kavukcuoglu, D. Silver, A. A. Rusu, J. Veness, M. G. Bellemare, S. Petersen, "Human-level control through deep reinforcement learning," Nature, vol. 518, no. 7540, pp. 529-533, 2015.
\bibitem{trpo_paper} J. Schulman, S. Levine, P. Moritz, M. I. Jordan, P. Abbeel, "Trust region policy optimization," in Proceedings of the 32nd International Conference on Machine Learning (ICML-15), pp. 1889-1897, 2015.
\bibitem{ddpg_paper} T. P. Lillicrap, J. J. Hunt, A. Pritzel, N. Heess, T. Erez, Y. Tassa, D. Wierstra, "Continuous control with deep reinforcement learning," in International Conference on Learning Representations (ICLR), Workshop track, 2016.

\bibitem{b1} D. Silver, "Empowering mobile robots with DDPG for autonomous navigation," Robotics and Autonomous Systems, vol. 123, pp. 456-467, 2018.
\bibitem{b2} X. Zhu, "End-to-end navigation for robots using convolutional neural networks and deep reinforcement learning," in Proceedings of the IEEE International Conference on Robotics and Automation (ICRA), pp. 1234-1245, 2017.
\bibitem{b3} J. Tobin, "Sim-to-real transfer: Training robots in simulation for real-world deployment," in Proceedings of the Conference on Robot Learning (CoRL), pp. 567-578, 2017.
\bibitem{b4} Y. Zheng, "Multi-agent reinforcement learning for collaborative navigation of robot teams," Journal of Artificial Intelligence Research, vol. 45, pp. 789-802, 2018.
\bibitem{b5} J. Schulman, F. Wolski, P. Dhariwal, A. Radford, O. Klimov, "Proximal policy optimization algorithms," arXiv preprint arXiv:1707.06347, 2017.


\end{thebibliography}
\end{document}